# Wood-leaf classification of tree point cloud based on intensity and geometrical information

**Author:** Jingqian Sun[1], Pei Wang[1*], Zhiyong Gao[2], Zichu Liu[1], Yaxin Li[1], Xiaozheng Gan[1]

**Address:** [1]School of Science, Beijing Forestry University, No.35 Qinghua East Road, Haidian District, Beijing 100083, China

[2]Research Institute of Petroleum Exploration and Development, Petrochina, Beijing, 100083, China.

**Corresponding author:** Pei Wang[1*]

**E-mail:** Jingqian Sun: sunjq_2019@bjfu.edu.cn

Pei Wang: wangpei@bjfu.edu.cn

Zhiyong Gao: gzy@petrochina.com.cn

Zichu Liu: lzc0330@bjfu.edu.cn

Yaxin Li: liyaxin_2018@bjfu.edu.cn

Xiaozheng Gan: ganxiaozheng@bjfu.edu.cn

[*]**For corresponding E-mail:**

wangpei@bjfu.edu.cn

# Abstract

Terrestrial laser scanning (TLS) can obtain tree point cloud with high precision and high density. Efficient classification of wood points and leaf points is essential to study tree structural parameters and ecological characteristics. By using both the intensity and spatial information, a three-step classification and verification method was proposed to achieve automated wood-leaf classification. Tree point cloud was classified into wood points and leaf points by using intensity threshold, neighborhood density and voxelization successively. Experiment was carried in Haidian Park, Beijing, and 24 trees were scanned by using the RIEGL VZ-400 scanner. The tree point clouds were processed by using the proposed method, whose classification results were compared with the manual classification results which were used as standard results. To evaluate the classification accuracy, three indicators were used in the experiment, which are Overall Accuracy (OA), Kappa coefficient (Kappa) and Matthews correlation coefficient (MCC). The ranges of OA, Kappa and MCC of the proposed method are from 0.9167 to 0.9872, from 0.7276 to 0.9191, and from 0.7544 to 0.9211 respectively. The average values of OA, Kappa and MCC are 0.9550, 0.8547 and 0.8627 respectively. Time cost of wood-leaf classification was also recorded to evaluate the algorithm efficiency. The average processing time are 1.4 seconds per million points. The results showed that the proposed method performed well automatically and quickly on wood-leaf classification based on the experimental dataset.

# 1. Introduction

Trees are very important in the ecological environment [1]. Trunks，branches and leaves are the main components of the above-ground biomass of trees [2]. Leaves are related to photosynthesis, respiration, transpiration, and carbon sequestration, while trunks composed of xylem and conduits are mainly used to transport water and nutrients. They have different physiological functions. Therefore, separating woody parts and leaves is the basis for some research, such as leaf area index (LAI) estimation, tree crown volume estimation, and diameter at breast height (DBH) estimation.

Laser scanning technology can be divided into three categories according to the platforms, which are spaceborne laser scanning, airborne laser scanning and terrestrial laser scanning (TLS) [3]. In forestry inventory, the spaceborne and airborne laser scanning are mainly used to obtain the information of large-scale forests to achieve the biomass estimation [4], species classification [5,6], tree height estimation [7], basal area estimation [8], carbon mapping [9], estimating forest structure [10]. Compared to the spaceborne and airborne laser scanning, TLS has the advantage of obtaining the trunk and branch information in detail from the viewpoint below the canopy. Therefore, tree point cloud can reflect the structural characteristics of trees better with less occlusion, and is a good complementary to other large-scale inventory method [3].

In recent years, TLS was widely used to obtain tree point cloud data which includes the woody part and the leaf part. Leaf point cloud is often used to estimate LAI [11-13], leaf area density [14,15], tree crown volume [16,17]. Similarly, wood point cloud is often used to calculate the parameters, such as tree position, DBH [18], tree branch and stem biomass [19,20], tree volume [21,22], stem curve [23,24]. They can be used together to calculate the gap fraction and the effective Plant Area Index [25,26] and tree biomass estimation [27,28]. Therefore, the wood-leaf classification is the basis of many forest inventory research. What's more, to some extent, the accuracy of wood-leaf classification affects the estimation accuracy of the above-mentioned parameters.

The intensity information is often obtained in the TLS data and is different for wood and leaves. Béland et al. achieved the wood-leaf classification by using the distance-based intensity normalization [29]. Some researchers used dual-wavelength LiDAR systems to realize wood-leaf classification based on the difference between the intensity of wood points and leaf points [30-32]. Zhao et al. used the intensity information of the multi-wavelength fluorescence LiDAR (MWFL) system to obtain the separation of vegetation stems and leaves [33]. However, the random and variable leaf positions and postures result in a wide distribution of the leaf points intensity which has an overlap with the distribution of wood points intensity. Therefore, it is hard to separate wood points and leaf points only by using an intensity threshold. The dual-wavelength system and the multi-wavelength system can improve the classification by using different thresholds in different wavelength respectively.

The geometric information and density information of tree point cloud data also were used to realize the wood-leaf classification. The skeleton points and k-dimensional tree (kd-tree) based on the geometric information of point cloud was used to classify wood points and leaf points [34]. Ma et al. proposed a method to separate photosynthetic and non-photosynthetic substances based on geometric information [35]. Ferrara et al. proposed a method to classify wood points and leaf points by using the Density-Based Spatial Clustering of Applications with Noise (DBSCAN) algorithm [36]. Xiang et al. used the skeleton points to classify plant stems and leaves [37]. Wang et al. utilized the recursive point cloud segmentation and regularization process to classify wood points and leaf points automatically based on the geometric information [38].

Some machine learning algorithms were also used to classify wood points and leaf points. Yun et al. used the semi-supervised support vector machine (SVM) to classify wood and leaves by extracting multiple features from point cloud data [39]. Zhu et al. classified wood and leaves by using a random forest (RF) algorithm [40]. Vicari et al. presented a new method combining unsupervised classification of geometric features and shortest path analysis to classify wood and leaf points [41]. Liu et al proposed different automated SVM classification methods for stem-leaf classification of potted plant point clouds [42] and wood-leaf classification of tree point clouds [43]. Krishna Moorthy et al. realized the wood-leaf classification by using radially bounded nearest neighbors on multiple spatial scales in a machine learning model [44]. Morel et al. proposed a method to classify wood points and leaf points based on deep learning and a class decision process [45]. Due to the laborious and time-consuming manual selection of training data in the establishment of the classifier, the automation and the efficiency of machine learning methods decreased.

This paper proposes a fast and automated wood-leaf classification method. A three-step classification is constructed to classify the points by using a intensity threshold at first, and then classify more points into leaf points by using KNN and voxel. Next, the wood points verification is fulfilled to correct some mis-classified wood points to improve the classification accuracy.

This paper is organized as follows: Section 2 described the experimental data used in the paper, and proposed the method and explained the method in detail. Section 3 demonstrated the separation result of tree point clouds. Section 4 analyzed the results and discussed the advantages and limits. Section 5 summarized the characteristics of the proposed method and looked forward to the future work.

## 2. Materials and Methods

### 2.1 Experimental data

The experiment data was collected in the Haidian Park, Haidian District, Beijing, China in June 2016. Three single scans were fulfilled by using the RIEGL VZ-400 TLS scanner which was manufactured by RIEGL company (RIEGL Laser Measurement

Systems GmbH, 3580 Horn, Austria). The characteristics of RIEGL VZ-400 scanner are listed in Table 1.

Table 1. The characteristics of RIEGL VZ-400 scanner.

| Technical parameters | |
|---|---|
| The farthest distance measurement | 600 m (natural object reflectivity≥90%) |
| The scanning rate (points / second) | 300000 (emission), 125000 (reception) |
| The vertical scanning range | -40° ～ 60° |
| the horizontal scanning range | 0° ～ 360° |
| Laser divergence | 0.3 mrad |
| The scanning accuracy | 3 mm (single measurement), 2 mm (multiple measurements) |
| The angular resolution | better than 0.0005° (in both vertical and horizontal directions) |

Some leaf-on plantation trees were scanned, and the obtained point cloud data contained the 3D data and intensity data. Twenty four willow trees (*Salix babylonica Linn and Salix matsudana Koidz*) were extracted from three single-scan scene point clouds manually by using the RISCAN PRO software. The 24 tree point clouds were demonstrated and numbered in Figure 1. The total tree heights (TTH) of these trees were from 8.82 m to 15.18 m, and their DBHs were from 14.2 cm to 29.3 cm.

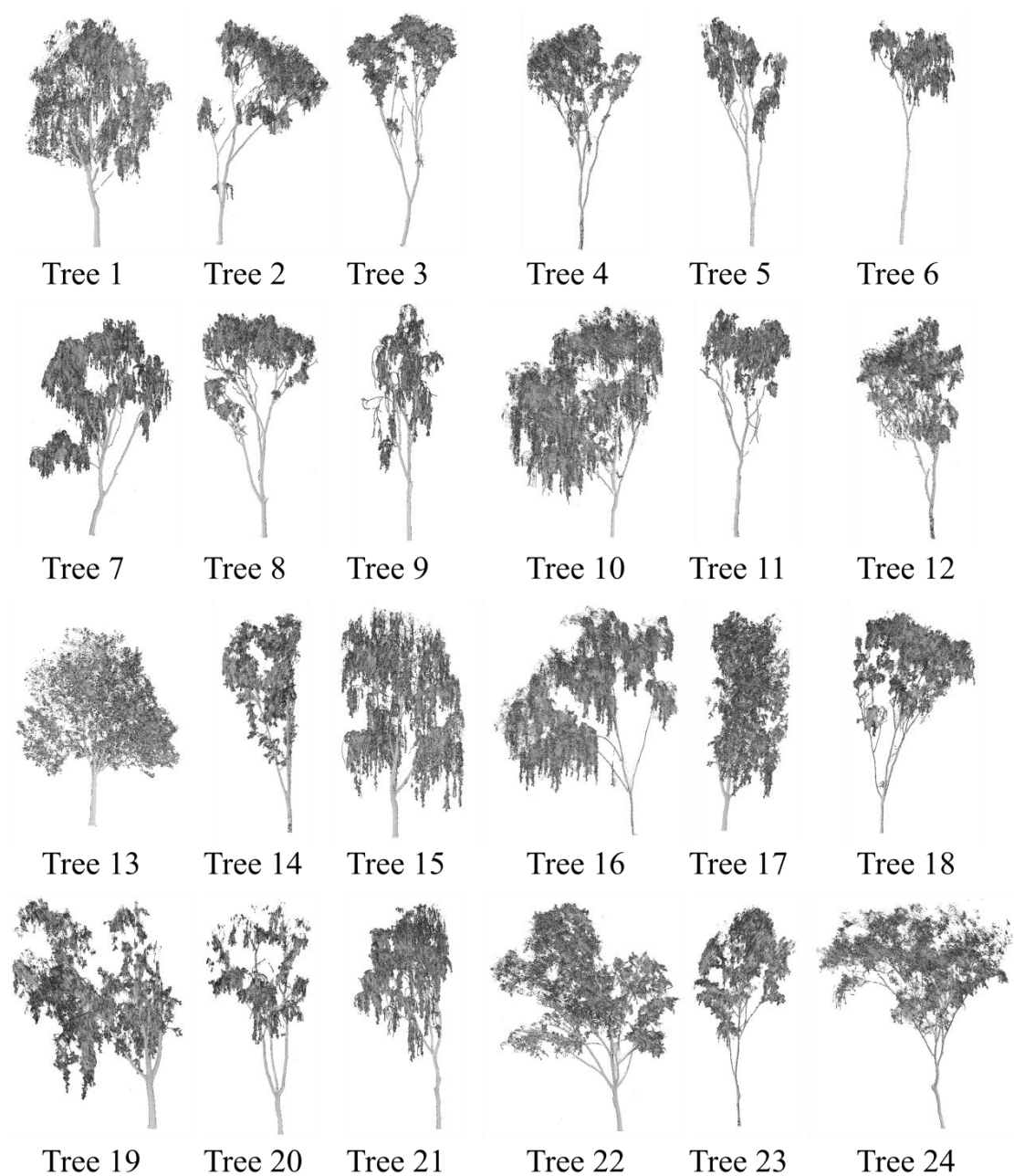

Figure 1. Extracted 24 tree point clouds.

In order to analyze the classification results and evaluate the proposed method, manual wood-leaf classification was performed for each extracted tree point cloud data in the CloudCompare software (An open source project is being defined by the GNU General Public License (GPL)). Wood points and leaf points of each tree are manually classified and the processing time of each tree is about 3-5 hours. The manual classification results were regarded as the standard classification results. Tree 5 was selected to demonstrate the typical manual classification result (Figure 2). The wood points were shown in brown color, and leaf points were shown in green color.

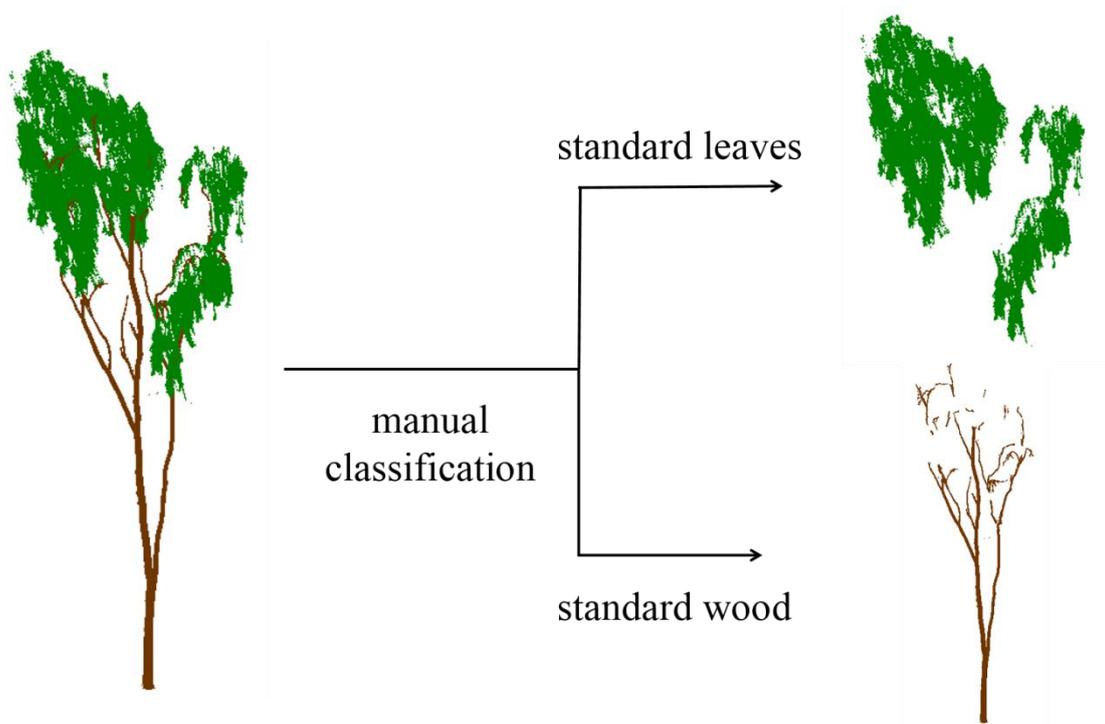

Figure 2. The manual standard classification result of tree 5.

**2.2 Method**

The proposed method aims to classify the wood points and leaf points automatically, whose full process is shown in detail in Figure 3. There are two main parts in the method. One part is the three-step classification based on the information of intensity, K nearest neighbors, point density in the voxel and voxel neighbors. The other part is the wood points verification which could correct part of the mis-classified points.

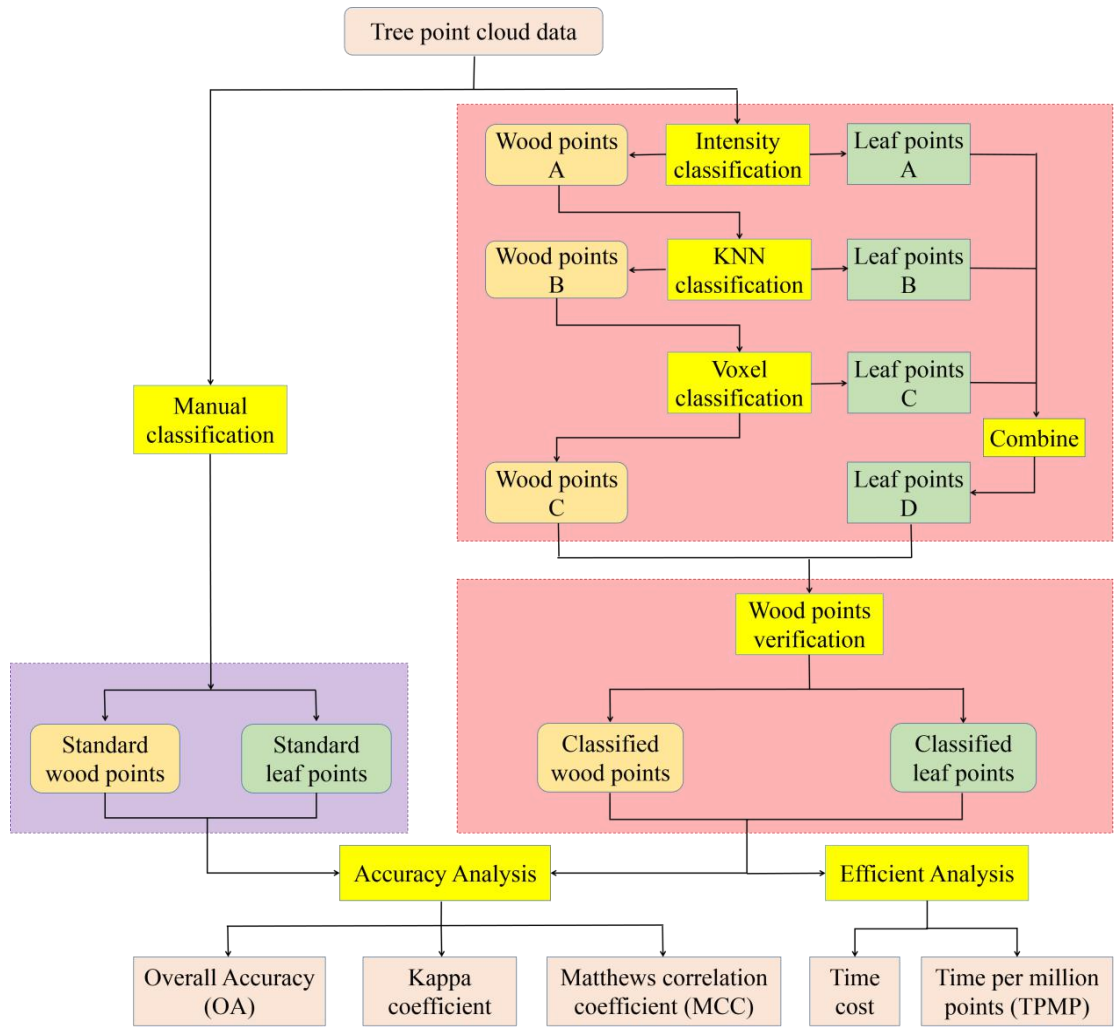

Figure 3. Flowchart of proposed method.

The above-mentioned three-step classification mainly focus on finding out the leaf points as much as possible.

First, intensity information is used to classify the tree points into wood points and leaf points. Trunk and branches are woody materials, and leaves are non-wood materials. They are different physical materials. Trunk and branches are hard and stable, while leaves are soft and often jittering by breeze. Therefore, the intensity values of trunk and branch points are greater than leaf points generally, while the intensity values of twig points almost have the similar magnitude as leaves. A suitable threshold of intensity can help us classify the raw tree points into wood points A and leaf points A, as shown in Figure 3.

Second, after intensity classification, some leaf points are still classified into wood points A because of their intensities. These leaf points are mostly sparse distributed in the 3D space which results in longer distances between the nearest neighbors than wood points. However, the real wood points mostly have shorter distances from their neighbor points than leaf points (as shown in Figure 4). Therefore, the K nearest neighbors are used to further find out these sparse distributed leaf points B in wood points A. The remained part of wood points A is wood points B.

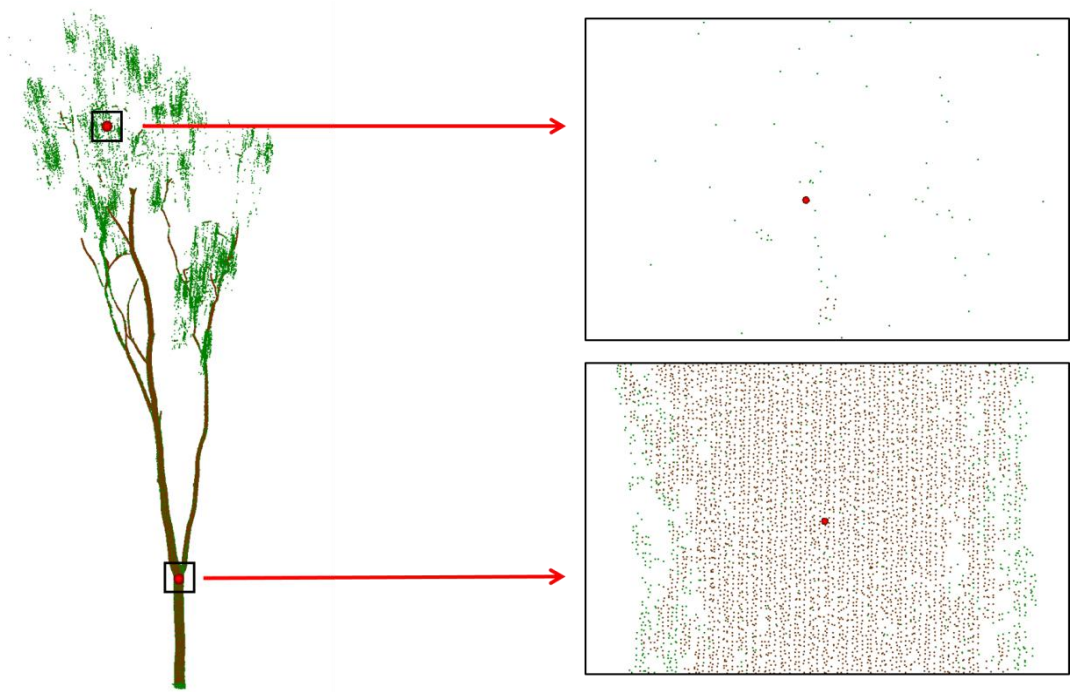

Figure 4. The spatial details of wood points and leaf points in wood points A.

Next, voxel was used to evaluate the point density on a larger scale. Considering the homogeneity and connectivity of trunk and branches, generally the points belonging to the same voxel are most likely to have the same properties. The whole point cloud space was divided into many voxels, and wood points B were classified as wood points C and leaf points C according to voxel features and neighbor relations.

After the above three-step classification processing, most leaf points were extracted from the total points. And the three parts of leaf points, leaf points A, leaf points B and leaf points C are combined into leaf points D. However, some wood points were still mis-classified as leaf points. These wood points usually are little bit far away from their neighbor wood points, and they can not meet the previous classification requirements of wood points. Therefore, voxel and intensity were both used to make a comprehensive verification to modify the categories of these points. At this phase, the number of wood points increases and the number of leaf points decreases.

At last, the classified wood and leaf points are ready to be evaluated. Three indicators on the accuracy and two indicators on the efficiency analysis will be given based on the classified results.

### 2.2.1. Intensity classification

As the first step of the three-step classification, the intensity classification would complete the rough classification of wood and leaf points. The generation of intensity information is complicated, which is related to the material, surface roughness, incident angle, measurement distance and object shape etc. [46]. Generally, as mentioned above, the wood points probably have greater intensity values than the leaf points under the same circumstances. However, there are some reasons which could decrease the intensity value of wood points or increase the intensity value of leaf points. For example, the wood surface is more rough than leaf surface, and some leaves may be face the scanner directly. Therefore, the intensity value distributions of wood and leaf points are partially overlapping. Most but not all points could be classified into wood and leaf points by using a simple threshold of intensity value. A small part of points are mis-classified because of their confusing intensity values.

The intensity threshold $I_t$ used in the method is generated adaptively for different tree point cloud. First, based on the principle of Random Sample Consensus (RANSAC), $n$ points in the tree point cloud were randomly selected as seed points. Second, the tree point cloud data was sampled spherically using the automatic random sampling method proposed in [42]. The spherical sampling took the seed points as the centers of the spheres and $r$ as the radius. Then, the sampling points in each sphere are projected onto the horizontal plane, and the projection density is calculated. The distribution of wood points is more concentrated than leaf points. Due to the difference between the spatial distributions of wood points and leaf points, their projection density would be different significantly. The projection density of wood points is larger than leaf points. Therefore, wood points and leaf points can be distinguished based on the projection density. After experimental tests, $n$ was selected as 1000, and $r$ was selected as 0.03m.

As shown in Figure 5, tree 5 point cloud was sampled with 1000 spheres, and the histogram of the projection density distribution of these spheres is plotted. According to previous assumptions, most spheres with high projection densities are more likely to contain wood points, while most spheres with low projection densities are more likely to contain leaf points. Therefore, the entire density interval [$\rho_{min}$, $\rho_{max}$] is quartered. The calculation of $\rho_{1/4}$ and $\rho_{3/4}$ is shown in equation 1. And the red and blue vertical lines in Figure 5 represent $\rho_{1/4}$ and $\rho_{3/4}$ respectively. The points contained in the sample spheres with density greater than $\rho_{3/4}$ are defined as wood points A in Figure 3; the points contained in the sample spheres with density less than $\rho_{1/4}$ are defined as leaf points A in Figure 3. The sampling results are shown in Figure 6, where the red points are the sampling leaves points and the blue points are sampling wood points.

$$\begin{cases} \rho_{1/4} = \rho_{min} + \dfrac{\rho_{max} - \rho_{min}}{4} \\ \rho_{3/4} = \rho_{max} - \dfrac{\rho_{max} - \rho_{min}}{4} \end{cases} \quad (1)$$

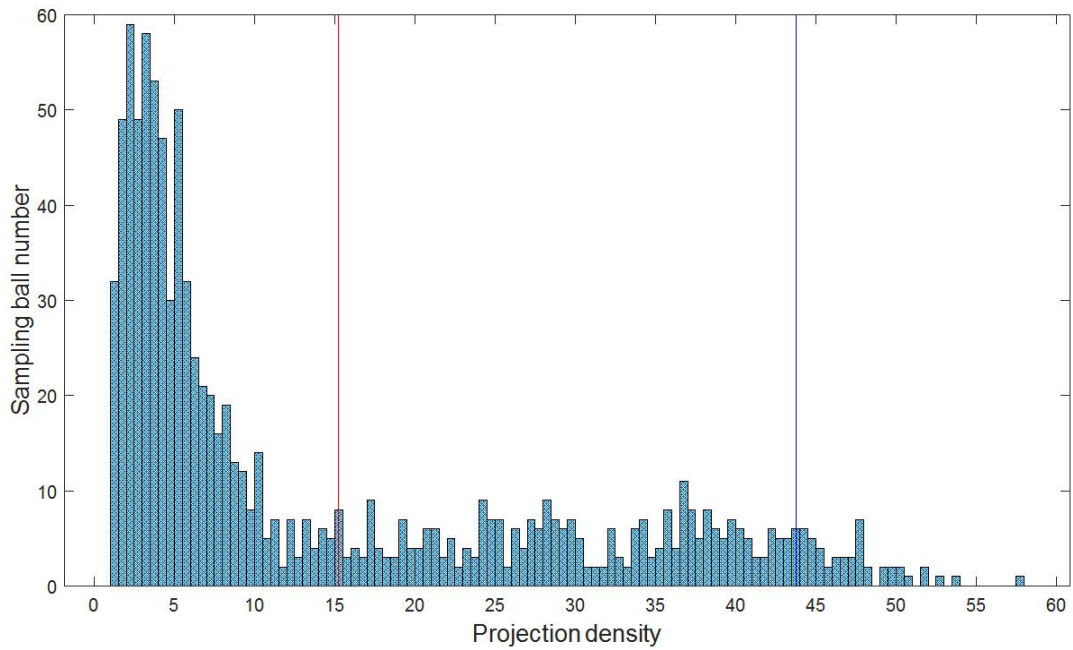

Figure 5. Histogram of the projection density distribution of randomly sampling spheres based on tree 5.

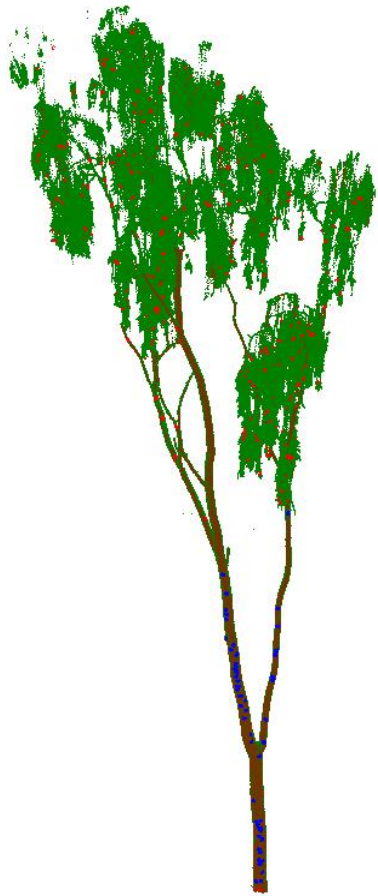

Figure 6. Display diagram of sampling results of tree 5.

Due to the theory of RANSAC, the sampled points can approximate express the intensity distribution of the original point cloud. Based on the wood-leaf classification results of the sampled points, the intensity was analyzed. Although the intensity values of the two parts have a relative concentrated distribution respectively, there still be overlapping areas with high probability. As shown in Figure 7, the intersection point of the wood and leaf points intensity distributions is used to separate the two parts. Most points can be classified into the correct class, while some points were classified wrongly.

The sampled and classified wood points and leaf points are used to fit the curves of their intensity distribution respectively. And the intersection point of these two fitted curves was calculated and used as the separation threshold $I_t$, which is plotted in Figure 7(b) with red color. The separation threshold $I_t$ is adaptive for each tree point cloud in the proposed method.

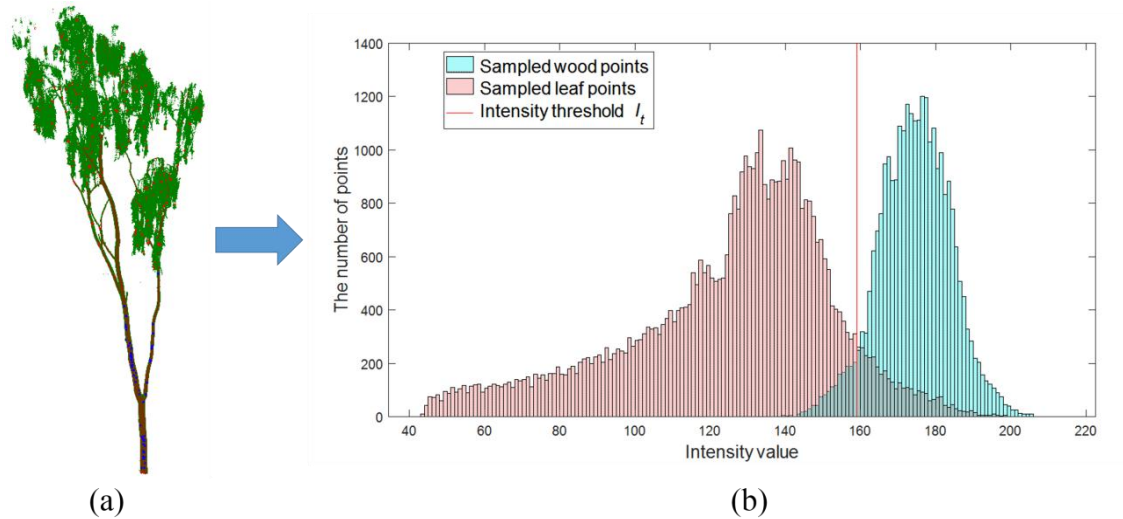

(a)          (b)

Figure 7. Demonstration of sampling results and intensity threshold based on tree 5.

**2.2.2. Neighborhood classification**

Further classification was needed because of the above coarse intensity classification. As shown in Figure 7, the overlaps of the intensity result in some points classified into wrong categories. The now classified wood points A and leaf points A are both composed of two parts. One part is the correct classified points, and the other part is the wrong classified points. The neighborhood classification was used to find the leaf points that are the wrong classified points in the wood points A.

The wood points are in more regular distribution comparing with the leaf points which scatter in the 3D space with different leaf postures. In a local area, the wood points are arranged closer to a plane, while the leaf points are more discrete than wood points. This is because the shape of the woody part of a tree is relatively stable in space, whereas the leaves themselves have a more chaotic distribution and may be affected by wind during data collection, which can cause jitter and further increase the dispersion of leaf points. Therefore, it can be inferred that the spatial distribution of real wood points in the wood points A were also compact and dense, and mis-classified leaf points are sparse and discrete. The degree of dispersion of leaf points would even increase after intensity classification because of the correct classification of most leaf points. Therefore, the KNN algorithm was considered to find some leaf points in the now wood points category further. In this paper, 8 nearest neighbors were used to separate the wood and leaf points.

We proposed to establish a kd-tree, and calculate the average distance $d_a$ between each point and its 8 nearest neighbors. If the point is on the woody part, the local area of the 8 nearest neighbors could be hypothesized as a small plane with a high probability, as shown in Figure 8(a).

First, calculate the spacing value $S_s$ of the target point at the range $\sigma$ with the angular step width $\theta_s$.

$$S_s = d_p \times \theta_s \qquad (2)$$

where $d_p$ is the distance value between the target point and the scanner.

Then, the $d_a$ on the plane can be calculate as following,

$$d_a = \frac{\sum_{n=1}^{8} d_n}{8} \qquad (3)$$

where $d_n$ is the distances from the target point to its 8 nearest neighbors.

As shown in Figure 8(a), the red point represents the target point, the $d_n$

value of the four yellow points is $S_s$, and the $d_n$ value of the four green points is $\sqrt{2}S_s$. Then the average distance $d_a$ can be calculated as $\frac{1+\sqrt{2}}{2}S_n$. Considering that the trunk and branches maybe inclined sometimes (Figure 8(b)), we assume the angle of inclination is no more than 45 degrees. So the maximum value of $d_a$ is about $1.71S_s$.

The $d_a$ of each point maybe different because of their different $\sigma$ value. Therefore, a ratio threshold $th_r$ for $d_a$ can used to classify the points into wood and leaf categories, which $th_r$ is 1.71. If the ratio of the target point is smaller than $th_r$, the point is classified into wood points category; otherwise, the point is classified into leaf points category.

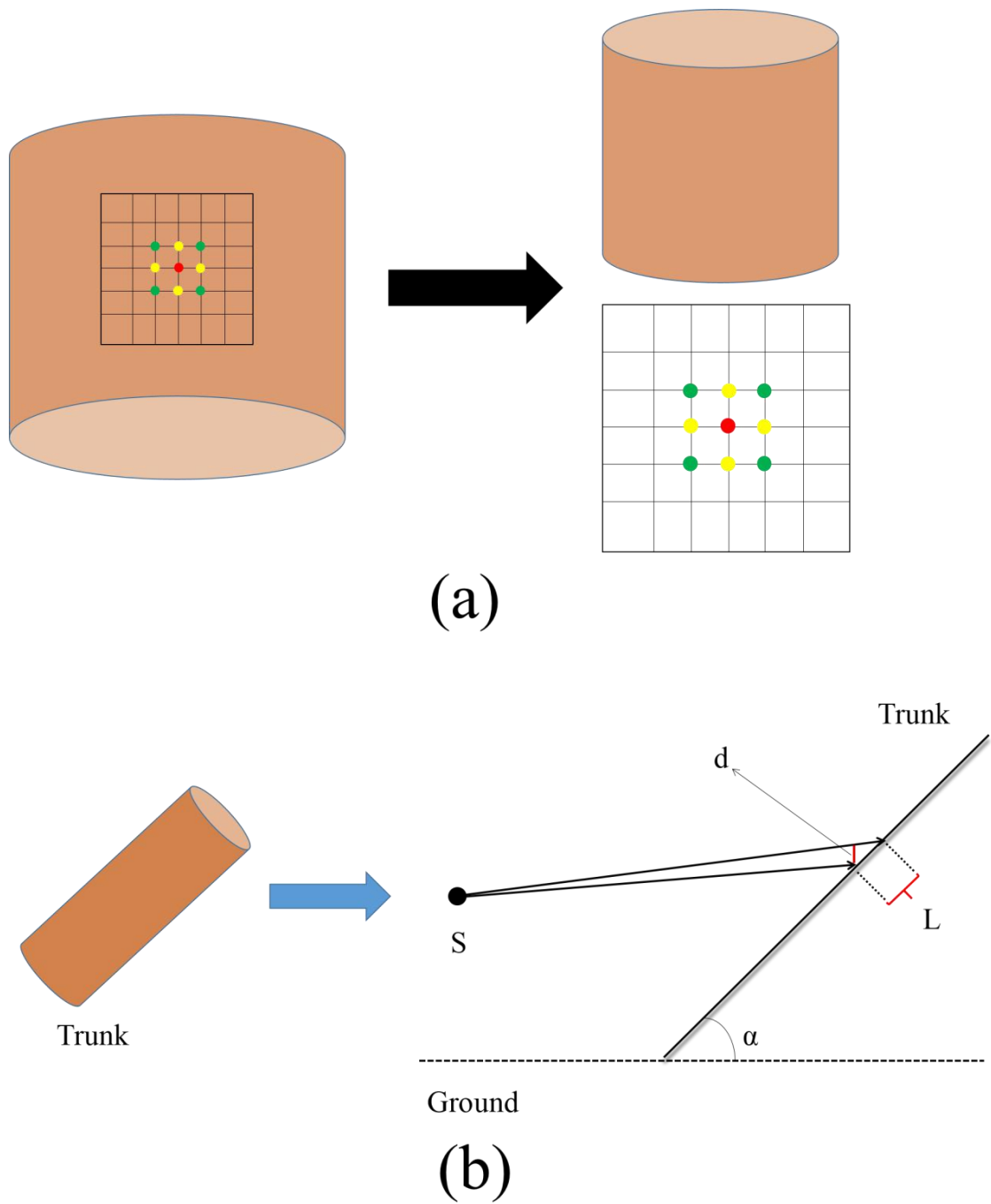

Figure 8. The demonstration of 8 nearest neighbors in the neighborhood

classification.

**2.2.3. Voxel classification**

By using the intensity information and neighbors information, most points are classified correctly. In this phase, more leaf points could be found out by using voxel classification described as follows.

The wood points B were measured in three dimensions (x,y,z). Each dimension was divided into 100 equal parts according to the specific dimensional information of points. That is to say, total 1,000,000 voxels cover the whole space containing the wood points B. Considering the homogeneity and connectivity of trunk and branches, generally the points belonging to the same voxel are most likely to have the same properties.

Now, there are still a few leaf points in the classified wood points B, and the remained leaf points are more sparse and discrete than ever. The fact led us to hypothesize that the voxels containing leaf points have small amount points in the voxels which result in smaller point densities. Therefore, the point density could be considered to determine the leaf points in wood points B.

However, the amount of points $Num_s$ should be contained in a voxel is affected by distance and angle. To simplify the determination, a ratio value $R$ was proposed to reduce the influence of distance and angle on point density in voxels.

First, $Num_s$ was calculated according to the distance $d_v$ and the angular step width $\theta_s$.

$$Num_s = \frac{Z_{size}}{d_v \times \theta_s} \times \frac{\sqrt{X_{size}^2 + Y_{size}^2}}{d_v \times \theta_s} \qquad (4)$$

where $X_{size}, Y_{size}, Z_{size}$ is the voxel sizes in three dimensions, $d_v$ is the distance between the center of voxel and the scanner.

Second, based on the actual number of points in the voxel $Num_r$, the ratio $R$ was calculated as following,

$$R = \frac{Num_r}{Num_s} \quad (5)$$

The ratio is smaller for the voxel mainly containing the leaf points, and larger for the voxel mainly containing the wood points. Taking tree 5 as an example (Figure 9), when the ratios of all voxels in wood points B were calculated, the histogram fitting curve (blue line in Figure 9(a)) and the derivative curve (blue line in Figure 9(b)) of the fitting curve are calculated. As shown in Figure 9(a), the fitting curve had a trend to going down. The location where the curve change trend went down significantly is selected as the ratio threshold which is 0.1 (a red vertical line). As shown in Figure 9(b), in this case, the derivative value is almost zero at $R = 0.1$. Therefore, in our method, it is hypothesized that a voxel is determined as a leaf voxel when the voxel ratio is smaller than 0.1; otherwise, the voxel is still determined as a wood voxel.

Additionally, considering the connectivity of trunk and branches, an isolated voxel would be determined as a leaf voxel even if it has a ratio greater than 0.1. And all points in the leaf voxel were classified into leaf points. Now, the wood points C and the leaf points C were obtained.

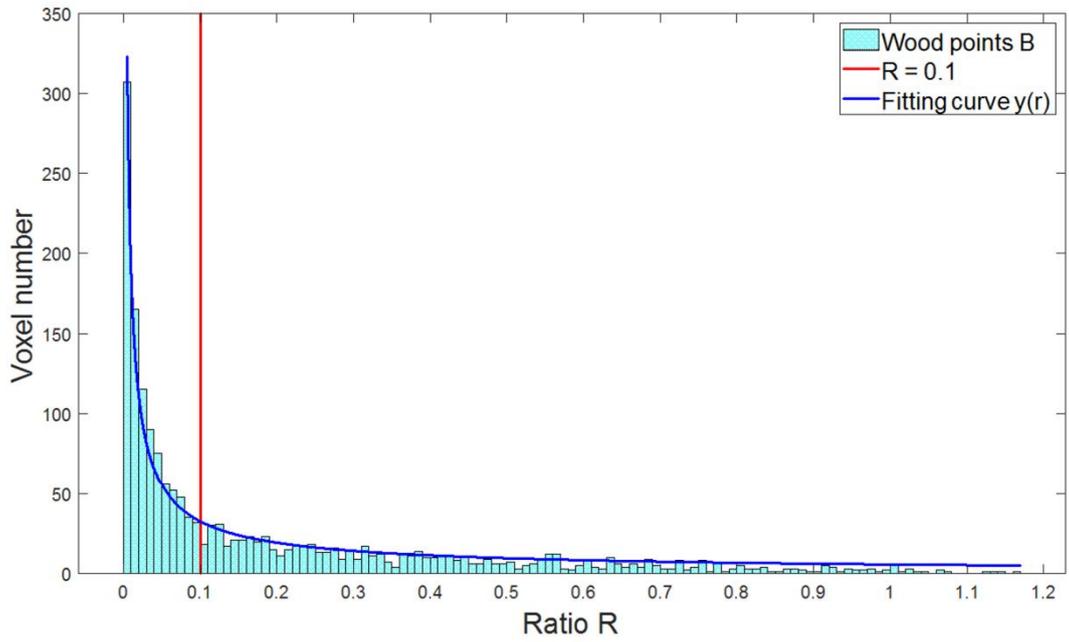

(a)

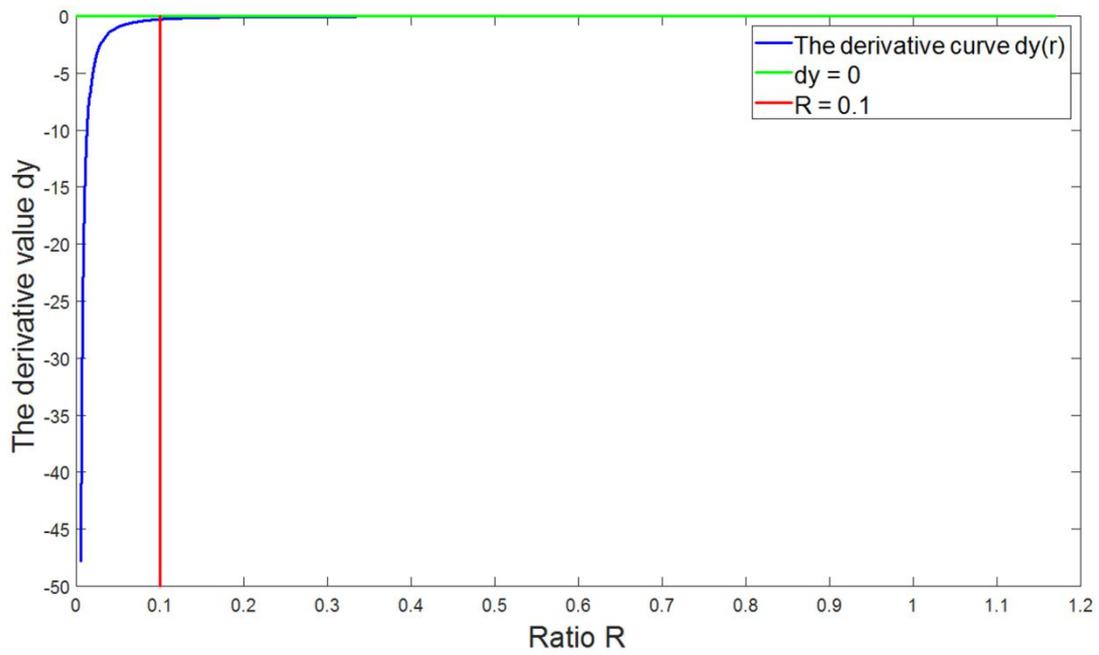

(b)

Figure 9. Demonstration of the threshold of voxel ratios.

**2.2.4. Wood points verification**

After the above-mentioned three-step classification operation, as many leaf points as possible have been found. The now leaf points D was composed of leaf points A, leaf points B and leaf points C. Meanwhile, a few wood points had been mis-classified in the process.

To further improve the classification accuracy, the voxel space constructed in the previous section was also used to verify the mis-classified wood points. For most experimental tree point clouds, there are generally fewer leaves in the lower part of the tree, and more in the upper part which are generally clustered close around the trunk. Therefore, the treatment of tree point cloud data was divided into two parts.

First, below 1/3 of the total tree height, the 3×3 voxel neighbors surrounding a wood voxel in the same voxel layer were checked. The neighbor voxel is determined as a new wood voxel if there are some points in it. The same process will happen to the new wood voxels until no more new wood voxel was found.

Second, above the 1/3 of the total tree height, the other procedure was taken to process the points. The 3×3×3 neighbor voxels of a wood voxel will be checked.

There are two different cases of mis-classified wood points. First, some wood points are mis-classified because their intensity values are smaller than the intensity threshold $I_t$. Second, some points are far away from the real wood points even though their intensity values is larger than $I_t$. To improve the above two cases, two variables $sd_1$ and $sd_2$ were introduced as the distance ratios. Among them, $sd_1$ is used to process the first case and $sd_2$ is used to process the second case. In the method, the values of $sd_1$ and $sd_2$ are 2 and 6, respectively.

(1). The $S_s$ of each wood point in the voxels is calculated based on equation 2;

(2). The distance $d_u$ between each wood point and leaf point in the voxels is calculated;

(3). Then, determine the new wood point according to the following equation,

$$\begin{cases} d_u \leq sd_1 * S_s & (a) \\ d_u \leq sd_2 * S_s \ \& \ P_i \geq I_t & (b) \end{cases} \quad (6)$$

Where $P_i$ is the intensity value of each leaf point. If a leaf point meets condition (a) or condition (b), the leaf point will be determined as a new wood point.

(4). Check each leaf point in the neighbor voxels to complete the new wood points verification;

(5). These new wood points were used in the above process until no more new wood points were found.

## 3. Results

### 3.1. Classification results

Taking tree 5 as an example, the process of the whole wood-leaf classification was demonstrated in Figure 10 which is corresponding to the outline of the method shown in Figure 3. It is clear that the points were classified into wood points and leaf points gradually by using the three-step classification, and improved by using the wood points verification.

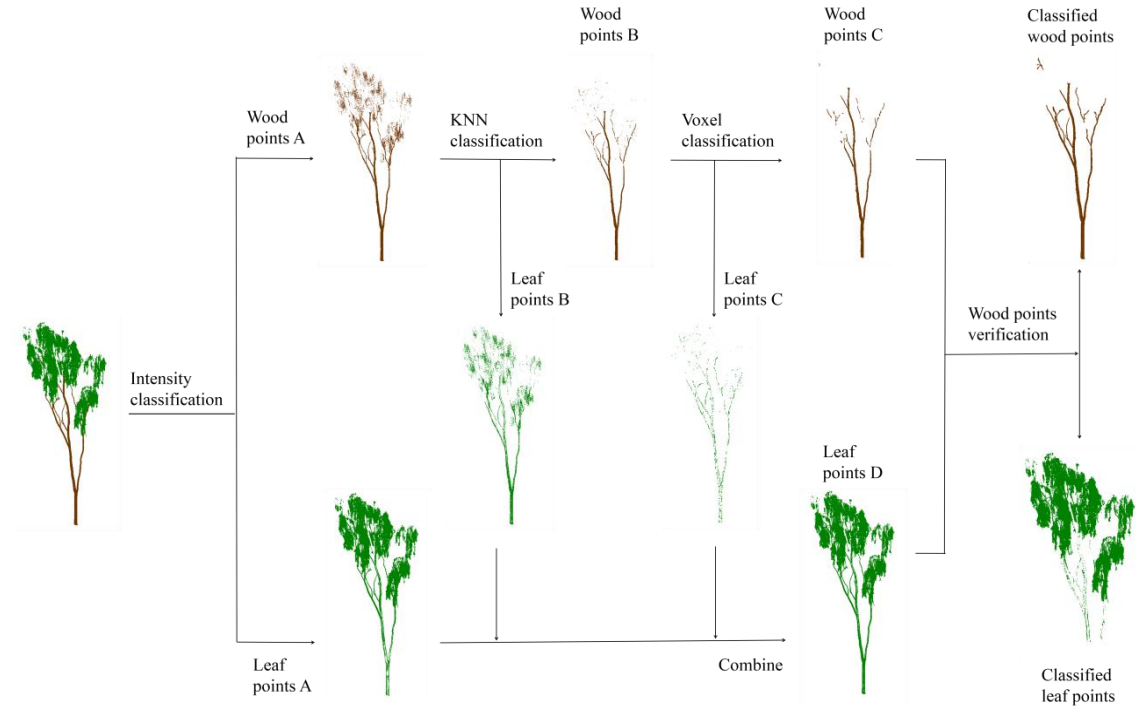

Figure 10. Schematic diagram of classification process of tree 5.

Table 2. The number of wood points and leaf points in the process of tree 5.

| Intensity classification | KNN classification | Voxel classification | | Wood points verification |
| --- | --- | --- | --- | --- |
| wood points A | wood points B | wood points C | | classified wood points |
| 301392 | 261408 | 242513 | | 393211 |
| leaf points A | leaf points B | leaf points C | leaf points D | classified leaf points |
| 763154 | 39984 | 18895 | 822033 | 671335 |

All the 24 tree point clouds were processed in the experiment. The classified wood points and leaf points of each tree were demonstrated in Figure 11. The wood points are colored in brown and the leaf points are colored in green. Obviously, the classification performed well on most of the tree point cloud.

The total points of each tree were listed in Table 3. As mentioned in the experimental data section, the manual classification results of all trees were used as the standards. And the amount of wood points and leaf points in the standard results were also listed in Table 3. What's more, the number of classified wood points and leaf points were provided too, including the number of true points and false points of each category.

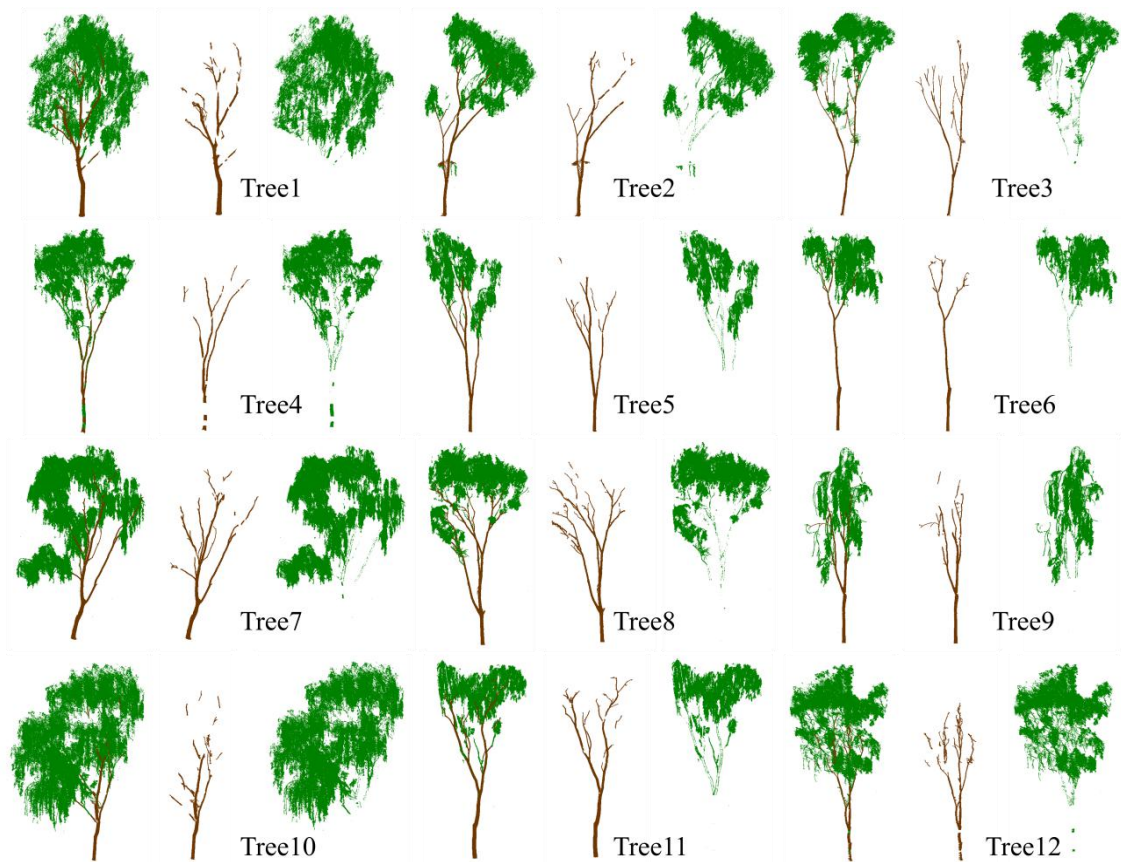

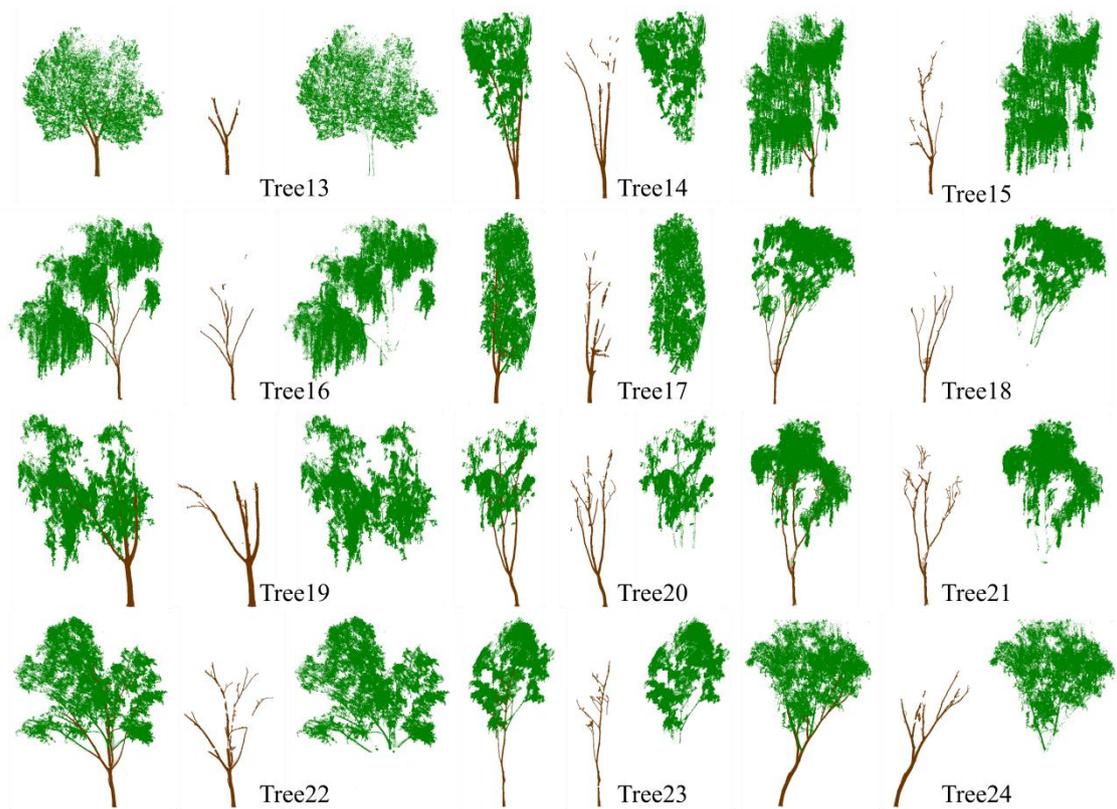

Figure 11. The demonstration of 24 tree classification results.

Table 3. The point statistics of 24 tree classification results.

| Tree / Number | Total points | Standard Results | | Classification Results | | | |
| --- | --- | --- | --- | --- | --- | --- | --- |
| | | Wood points | Leaf points | Wood points | | Leaf points | |
| | | | | True wood | False wood | True leaf | False leaf |
| 1 | 876657 | 150479 | 726178 | 128879 | 1215 | 724963 | 21600 |
| 2 | 716701 | 154548 | 562153 | 133791 | 5647 | 556506 | 20757 |
| 3 | 629250 | 190793 | 438457 | 166616 | 2080 | 436377 | 24177 |
| 4 | 733233 | 169071 | 564162 | 116880 | 651 | 563511 | 52191 |
| 5 | 1064546 | 427139 | 637407 | 384086 | 1592 | 635815 | 43053 |
| 6 | 971915 | 246251 | 725664 | 213843 | 1899 | 723765 | 32408 |
| 7 | 3398859 | 719573 | 2679286 | 638655 | 7436 | 2671850 | 80918 |
| 8 | 1162123 | 312819 | 849304 | 271612 | 4924 | 844380 | 41207 |
| 9 | 1068644 | 374865 | 693779 | 289835 | 3926 | 689853 | 85030 |
| 10 | 1210685 | 143532 | 1067153 | 105130 | 1653 | 1065500 | 38402 |
| 11 | 1318700 | 562884 | 755816 | 508514 | 1065 | 754751 | 54370 |
| 12 | 742280 | 193707 | 548573 | 140832 | 1491 | 547082 | 52875 |
| 13 | 203303 | 13301 | 190002 | 8801 | 37 | 189965 | 4500 |
| 14 | 1896619 | 482532 | 1414087 | 420063 | 7086 | 1407001 | 62469 |
| 15 | 1080397 | 109269 | 971128 | 88755 | 1962 | 969166 | 20514 |
| 16 | 980776 | 79224 | 901552 | 66944 | 184 | 901368 | 12280 |
| 17 | 841575 | 100118 | 741457 | 76668 | 8182 | 733275 | 23450 |
| 18 | 1357196 | 375669 | 981527 | 286918 | 4034 | 977493 | 88751 |
| 19 | 4925230 | 1329062 | 3596168 | 1128847 | 8731 | 3587437 | 200215 |
| 20 | 1716488 | 727900 | 988588 | 644566 | 6718 | 981870 | 83334 |
| 21 | 1275620 | 215761 | 1059859 | 179962 | 4550 | 1055309 | 35799 |
| 22 | 1301100 | 240684 | 1060416 | 150458 | 1391 | 1059025 | 90226 |
| 23 | 1315914 | 364161 | 951753 | 279447 | 3560 | 948193 | 84714 |
| 24 | 771395 | 165762 | 605623 | 118643 | 1805 | 603828 | 47119 |

### 3.2. Accuracy and efficiency analysis

Based on the results listed above, three indicators were used to access the classification accuracy by comparing with the standard results. N is the total number of tree points as following,

$$N = TP + FP + TN + FN \qquad (7)$$

Among them, TP indicates the number of correctly classified leaf points, TN indicates the number of successfully marked wood points, FP means the number of wood points that are incorrectly classified as leaf points, and FN describes the number of leaf points which are wrongly recognized as wood points.

The first indicator is OA which ranges from 0 to 1 and represents the probability that the overall classification is correct. However, OA performs not very well when the dataset is unbalanced. It is calculated by using the following equation,

$$OA = \frac{TP+TN}{N} \tag{8}$$

The second indicator is Kappa which is often used for consistency testing and can also be used to access the effect of classification. For the better performance on evaluating the classification of unbalanced dataset, Kappa coefficient is widely used for classification accuracy evaluation. The calculation result of Kappa coefficient is -1~1, but usually it falls between 0 and 1. Among them, 0~0.2 usually means very low consistency, 0.21~0.4 usually means general consistency, 0.41~0.6 usually means medium consistency, 0.61~0.8 usually means high consistency, 0.81~1 means almost complete consistency. And Kappa coefficient can be given by,

$$Kappa = \frac{p_o - p_e}{1 - p_o} \tag{9}$$

Where $p_o = \frac{TP+TN}{N}$, $p_e = \frac{(TP+FP)\times(TP+FN)+(TN+FN)\times(TN+FP)}{N \times N}$.

The third indicator is MCC which is similar to the Kappa coefficient and is also often used to measure the accuracy of classification [47]. Both MCC and Kappa are accuracy indicators that are used to evaluate the classification accuracy of unbalanced datasets. Some researchers believe MCC is better than Kappa coefficient on evaluating the classification accuracy [48]. Therefore, this paper uses both of these accuracy indicators to increase the scientificity and reliability of the research results. MCC value ranges from -1 to 1, where 1 means perfect prediction, 0 means no better than random prediction, -1 means complete inconsistency between prediction and observation. MCC can be calculated as follows:

$$MCC = \frac{TP \times TN - FP \times FN}{\sqrt{(TP+FP) \times (TP+FN) \times (TN+FN) \times (TN+FP)}} \quad (10)$$

All the three indicators of each tree were calculated and listed in following Table 4. As shown, the OAs of all the 24 trees are very high. The OAs of 24 trees range from 0.9167 to 0.9872, and the average OA value was 0.9550; the Kappa coefficient ranged from 0.7276 to 0.9191, and the average value was 0.8547; the MCC ranged from 0.7544 to 0.9211, and the average value was 0.8627. And there is almost no difference between Kappa and MCC.

Table 4. The results of accuracy and efficiency analysis.

| Tree / Number | Accuracy analysis | | | Time analysis | |
|---|---|---|---|---|---|
| | OA | Kappa | MCC | Time cost / ms | TPMP / ms |
| 1 | 0.9739 | 0.9032 | 0.9066 | 935 | 1067 |
| 2 | 0.9631 | 0.8870 | 0.8889 | 930 | 1298 |
| 3 | 0.9582 | 0.8979 | 0.9012 | 870 | 1383 |
| 4 | 0.9279 | 0.7726 | 0.7923 | 912 | 1244 |
| 5 | 0.9580 | 0.9113 | 0.9144 | 1901 | 1786 |
| 6 | 0.9647 | 0.9027 | 0.9061 | 1350 | 1390 |
| 7 | 0.9740 | 0.9191 | 0.9211 | 5547 | 1633 |
| 8 | 0.9603 | 0.8952 | 0.8983 | 1565 | 1347 |
| 9 | 0.9167 | 0.8076 | 0.8203 | 1625 | 1521 |
| 10 | 0.9669 | 0.8219 | 0.8331 | 1103 | 912 |
| 11 | 0.9579 | 0.9130 | 0.9162 | 2456 | 1863 |
| 12 | 0.9267 | 0.7923 | 0.8080 | 917 | 1236 |
| 13 | 0.9776 | 0.7837 | 0.8021 | 506 | 2489 |
| 14 | 0.9633 | 0.8995 | 0.9024 | 2981 | 1572 |
| 15 | 0.9792 | 0.8762 | 0.8808 | 990 | 917 |
| 16 | 0.9872 | 0.9080 | 0.9116 | 880 | 898 |
| 17 | 0.9624 | 0.8080 | 0.8115 | 791 | 940 |
| 18 | 0.9316 | 0.8164 | 0.8281 | 1789 | 1319 |
| 19 | 0.9575 | 0.8872 | 0.8919 | 12753 | 2590 |
| 20 | 0.9475 | 0.8910 | 0.8949 | 3517 | 2049 |
| 21 | 0.9683 | 0.8805 | 0.8843 | 1334 | 1046 |
| 22 | 0.9295 | 0.7276 | 0.7544 | 1392 | 1070 |
| 23 | 0.9329 | 0.8200 | 0.8315 | 1778 | 1352 |
| 24 | 0.9365 | 0.7913 | 0.8065 | 938 | 1216 |
| Mean | 0.9550 | 0.8547 | 0.8627 | / | 1423 |

The OA, Kappa and MCC of each tree are also plotted in Figure 12. Obviously, the overall classification accuracy evaluation given by OA is higher than Kappa and MCC. The plotted Kappa and MCC are almost the same. Especially, the OA values of trees 4, 12, 13, 22 and 24 are larger than 0.9, but their Kappa values are smaller than 0.8.

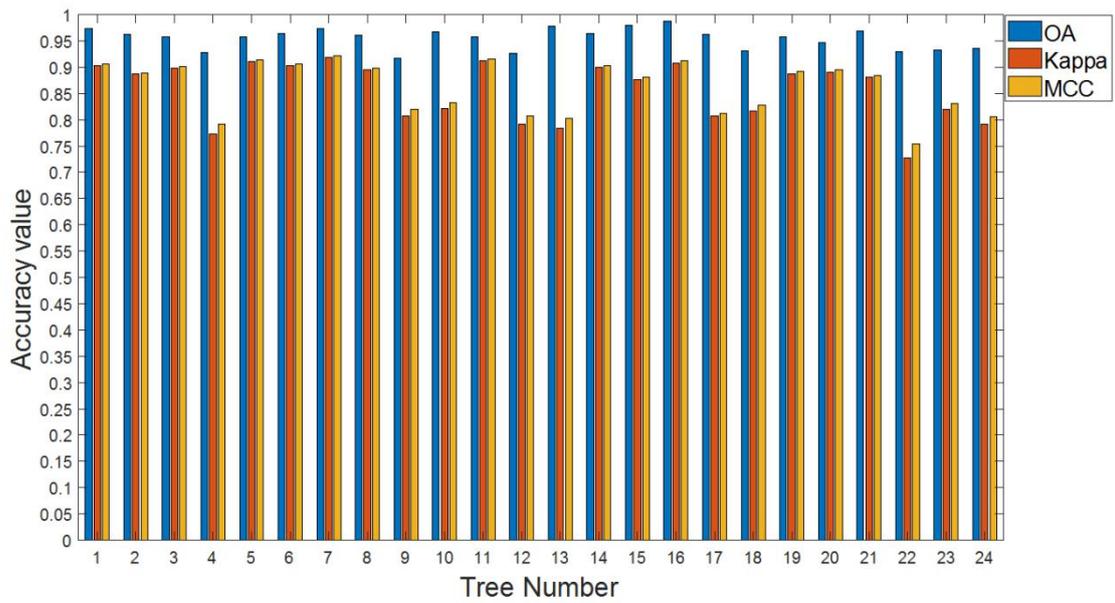

Figure 12. The histogram of OA, Kappa, MCC.

In terms of processing speed analysis, the classification time cost of each tree was recorded in Table 4. Because of the number of tree points are different, the time cost per million points also calculated and listed in Table 4. Generally, the more points, the more time the processing takes. As shown, most tree point clouds with less than 2 million points can be classified in 2s. Some trees cost a little bit more time considering their points. The time cost of each tree is shown and a curve is fitted based on the number of tree points (Figure 13).

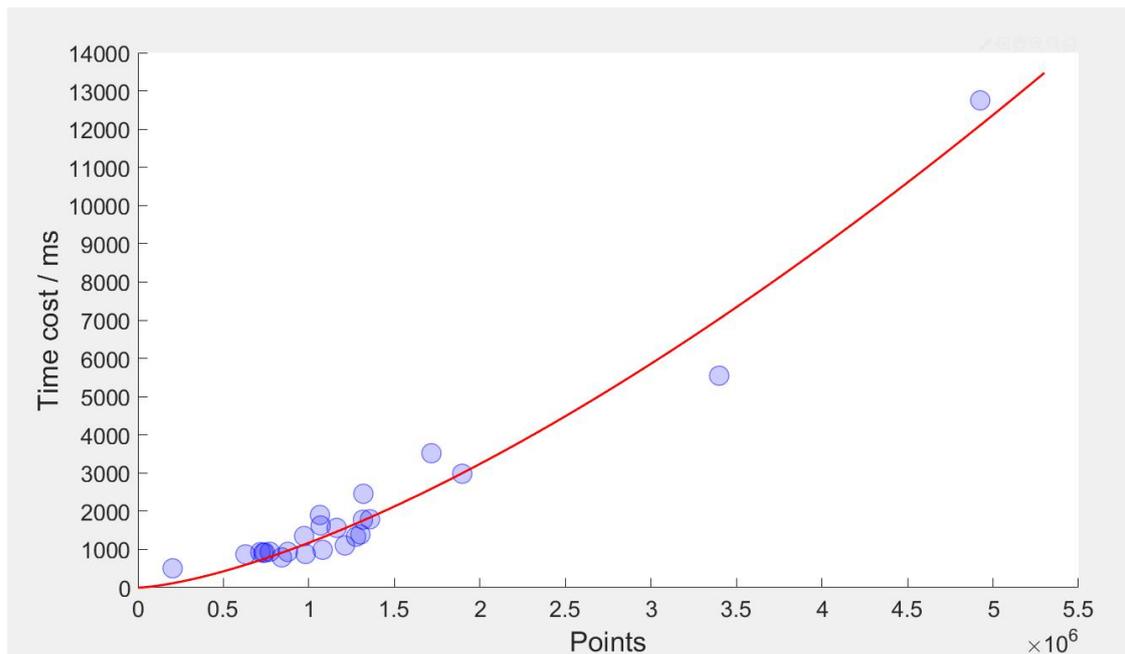

Figure 13. The relationship between the time cost and the number of tree points.

## 4. Discussion

In term of classification accuracy, as above mentioned, the proposed method had a good accuracy performance on the experimental dataset. Some reported classification results also performed well on their experimental dataset. Tao et al. processed three trees including two real trees and one simulation tree, and the corresponding Kappa coefficients were 0.79, 0.80 and 0.89, respectively [34]. Yun et al. treated 5 trees, and the OA ranged from 0.8913 to 0.9349 [39]. Zhu et al. processed 10 trees and achieved an average OA of 0.844, an average Kappa value of 0.75 [40]. Ferrara et al. processed 7 cork oak trees, and the OA values varied from 0.95 to 0.98, and the MCC values ranged from 0.76 to 0.88, and the Kappa coefficient ranged from 0.75 to 0.88 [36]. Vicari et al. processed a total of 10 filed tree point clouds, with OA ranging from 0.85 to 0.93, and Kappa coefficient ranged from 0.48 to 0.81 [41]. Krishna Moorthy et al.

processed 9 filed tree data, the classification accuracy ranged from 0.72 to 0.92, and the average accuracy was 0.876 [44]. Liu et al. processed 10 trees, and the OA ranged from 0.8961 to 0.9590, the Kappa coefficients varied from 0.7381 to 0.8755, and the means of OA and Kappa coefficient were 0.9305 and 0.7904 respectively [43]. Wang et al. processed 61 tropical trees, and the overall classification accuracy was $0.91 \pm 0.03$ in average [38]. Using the same dataset as [38], Morel et al. processed 35 trees in the dataset and used another indicator IoU (Intersection over Union) to evaluate the classification accuracy which ranged from 0.85 to 0.97 [45]. Morel et al. believed that manual classification results should not be used as standard results due to the lack of precise ground truth.

Besides the proposed method in the paper, most of the above reported methods cannot be compare directly because of their different experimental dataset. Therefore, the reported classification accuracies can only be used as references.

In terms of efficiency, the proposed method decreased the time cost because of automated processing. According to references, the classification efficiency was reported in literatures [38] and [41]. As reported, 90 seconds on average were needed to process each million points by using Wang's method. And Vicari introduced that the average processing time of each tree was 10 minutes without mentioning the amount of points. As shown in Table 4, we calculated the time cost of each tree in the experiment. The time cost of 24 trees range from 0.506 seconds to 12.753 seconds, and the average processing time is 1.4 second per million points. The efficiency of the proposed method is better than the other two reported methods.

As shown in Figure 11 and Table 4, some trees have good OA, Kappa and MCC values, such as tree 3, tree 5, tree11 and tree 16. However, there are some small twigs were classified into the leaf parts. Even if there are these misclassified twigs, the reduction of classification accuracy is little because of the small amount of points. According to the rules of the proposed method, these twigs are closer to the leaf points in intensity value and spatial distribution. This can be a future work for improving the proposed method, although there may not result in a big improvement in classification accuracy.

Some trees, such as tree 4, tree 12, tree13, tree 22 and tree 24, have larger OA values than 0.9, but smaller Kappa values than 0.8. To analyze the reasons, the intensity distributions of manual classified wood points and leaf points of each tree were plotted in Figure 14, in which the red line means the adaptive threshold.

Obviously, as shown in Figure 14, the number of leaf points is much larger than the wood points. And some threshold values are accurate, and some are not, which means that the random sampling strategy did not accurately reflect the intensity distribution when the number of leaf points and wood points differ greatly. Meanwhile, the overlapping areas account for a large proportion of the wood points, and the threshold is close to the peak of the intensity distribution of wood points. We believe that the disparity in the number of wood and leaf points, and the threshold of deviation lead to worse intensity classification which separate most points in tree point cloud. Therefore, the proposed method didn't obtain good performance on these tree point cloud.

All in all, although the characteristics of automation, high accuracy, and high speed in the experiment has been shown, the proposed method still need to be improved to several areas that can be improved to increase robustness.

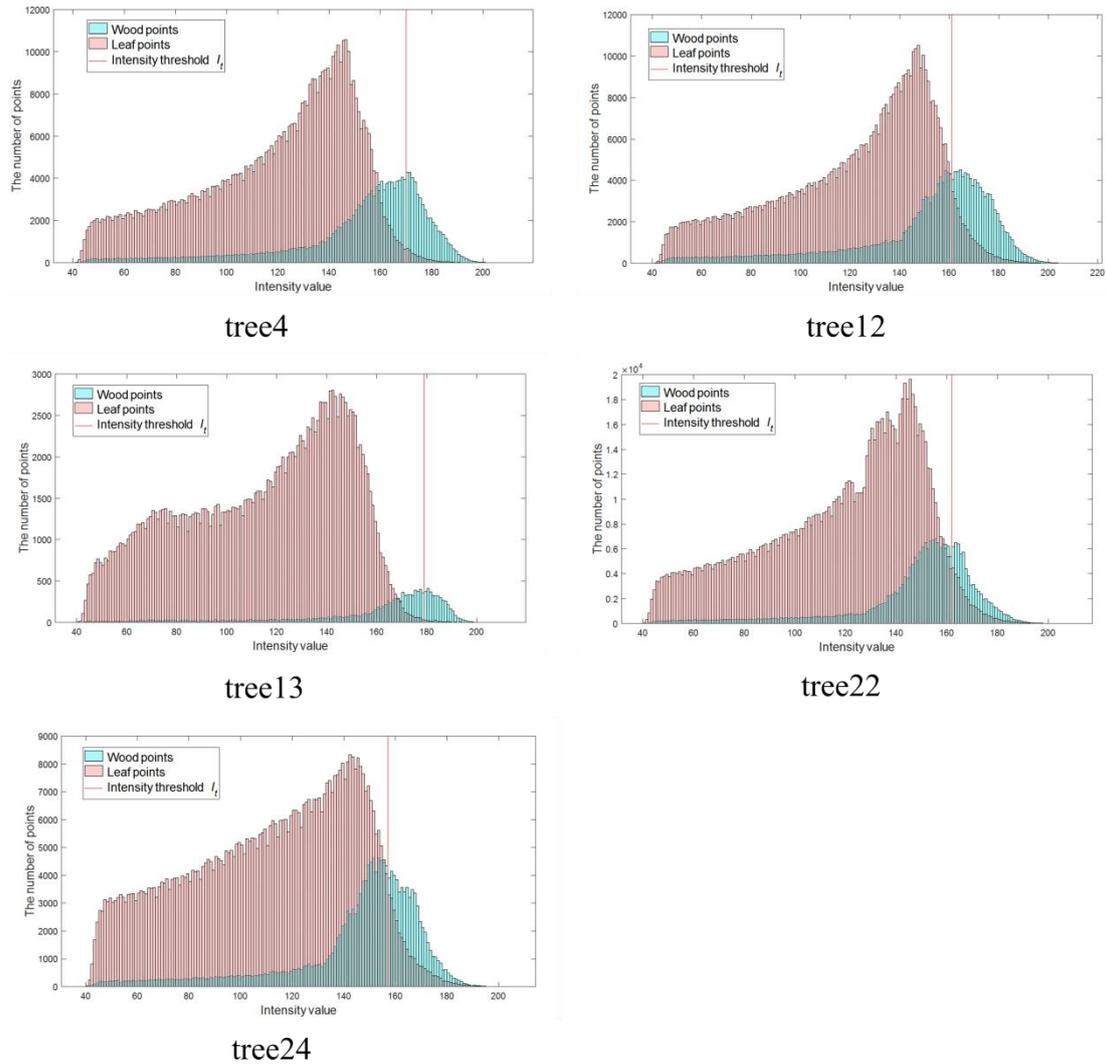

Figure 14. Display of intensity distributions for manual separation results of five trees and its adaptive intensity thresholds.

## 5. Conclusion

The paper proposed an automated wood-leaf classification method for tree point cloud using the intensity information and spatial information. The experiment shows that the proposed method is automated, accurate and high-speed. Even though the accuracy of the proposed method will be reduced when classifying some tree point clouds with the characteristics analyzed in discussion, the classification accuracy is still good.

The proposed method has good practical value and application prospects. In the future work, more trees will be tested to improve the accuracy and robustness of the method.

**Authors' contributions:** Conceptualization, Jingqian Sun and Pei Wang; Data curation, Zhiyong Gao, Zichu Liu, Yaxin Li and Xiaozheng Gan; Investigation, Jingqian Sun and Pei Wang; Methodology, Jingqian Sun and Pei Wang; Validation, Jingqian Sun; Writing – original draft, Jingqian Sun; Writing – review & editing, Pei Wang.

**Funding:** This research was funded by the Fundamental Research Funds for the Central Universities (No.2021ZY92).

**Conflicts of Interest:** The authors declare no conflict of interest.